\documentclass{ifacconf}

\usepackage{graphicx}      %
\usepackage{xcolor}
\usepackage{natbib}        %
\usepackage{mathtools}
\usepackage{nicefrac}
\usepackage{amsmath,amsfonts,amssymb}

\newcommand{\defeq}{\doteq}
\newtheorem{assumption}{\bf A\!\!}
\newcommand\norm[1]{\lVert#1\rVert}
\newcommand{\newpart}{}
\newcommand{\Ker}{K}
\newcommand{\RR}{\mathbb{R}}

\def \RKHS{\mathcal{H}}

\def\argmin{\operatornamewithlimits{arg\,min}}
\def \CH{\mathcal{H}}
\def \CD{\mathcal{D}}
\newcommand{\tr}{^\mathrm{T}}
\newtheorem{remark}{Remark}
\newcommand\eqind{\mathrel{\stackrel{\makebox[0pt]{\mbox{\normalfont\tiny d}}}{=}}}

\begin{document}
\begin{frontmatter}

\title{Improving Kernel-Based Nonasymptotic Simultaneous Confidence Bands\hspace{-0.3mm}\thanksref{footnoteinfo}}

\thanks[footnoteinfo]{This research was supported by the European Union within the framework of the Artificial Intelligence National Laboratory, RRF-2.3.1-21-2022-00004; and by the TKP2021-NKTA-01 grant of the National Research, Development and Innovation Office, Hungary.}

\author[First,Second]{Balázs Csanád Csáji}
\author[First,Third]{\qquad Bálint Horváth}

	\address[First]{Institute for Computer Science and Control (SZTAKI),\\
		E\"otv\"os Lor\'and Research Network (ELKH),\\
		13-17 Kende utca, H-1111, Budapest, Hungary}
	\address[Second]{Institute of Mathematics, E\"otv\"os Lor\'and University (ELTE),\\
	1/C Pázmány Péter sétány, H-1117, Budapest, Hungary}
    \address[Third]{Institute of Mathematics,  Budapest University of Technology and Economics (BME),
    1 Egry József utca, H-1111, Budapest, Hungary\\[1mm]	
	emails: \{horvath.balint, csaji.balazs\}@sztaki.hu}

\begin{abstract}                %
The paper studies the problem of constructing nonparametric simultaneous confidence bands with nonasymptotic and distribition-free guarantees. The target function is assumed to be band-limited and the approach is based on the theory of Paley-Wiener reproducing kernel Hilbert spaces. The starting point of the paper is a recently developed algorithm to which we propose three types of improvements. First, we relax the assumptions on the noises by replacing the symmetricity assumption with a weaker distributional invariance principle. Then, we propose a more efficient way to estimate the norm of the target function, and finally we enhance the construction of the confidence bands by tightening the constraints of the underlying convex optimization problems. The refinements are also illustrated through numerical experiments.
\end{abstract}


\end{frontmatter}
\vspace*{-5mm}
\section{Introduction}
\vspace*{-1mm}
One of the core problems of system identification, machine learning and statistics is {\em regression}, i.e., how to construct models from a sample of noisy input-output data.
The main task of regression is typically to estimate, based on a finite number of observations, the \textit{regression function}, which for a given input encodes the conditional expectation of the corresponding output \citep{cucker2007learning}. 

There are a number of well-known approaches to solve regression problems, such as least squares (linear regression), prediction error and instrumental variable methods, neural networks, and kernel machines \citep{gyorfi2002distribution}.

Standard approaches to regression often provide point estimates, while {\em region estimates}, which are vital for robust approaches and risk management, are typically constructed using the asymptotic distribution of the (scaled) estimation errors. On the other hand, from a practical point a view, methods with {\em nonasymptotic} and {\em distribution-free} guarantees are preferable. There are various types of region estimates that we can consider, which include confidence regions in the parameter space 
\citep{csaji2014sign}, confidence or credible bands for the expected outputs at given query points
\citep{Rasmussen2006},
and prediction regions for the next (noisy) observations \citep{vovk2005algorithmic,
garatti2019class}.

This paper focuses on building {\em simultaneous} confidence bands for the regression function. In a parametric setting such regions are simply induced by confidence regions in the parameter space, however, in a {\em nonparametric} setting these indirect approaches are typically not suitable. 

When the data are Gaussian, an impressive framework is offered by Gaussian process regression 
\citep{Rasmussen2006}, which can provide prediction regions for the outputs, and credible regions for the expected outputs. However, in practical situations the Gaussianity assumption is sometimes too strong, which motivates alternative approaches with weaker statistical assumptions.

In a recent paper a novel {\em nonasymptotic} method was suggested to build data-driven confidence bands for bounded, band-limited (regression) functions based on the theory of Paley-Wiener kernels \citep{csaji2022nonparametric}. It is {\em distribution-free} in the sense that only mild statistical assumptions are required about the noise on the observations, such as they are symmetric, independent from the inputs, and that the sample contains independent and identically distributed (i.i.d.) input-output pairs. On the other hand, the distribution of the inputs is assumed to be known, in particular, uniformly distributed.
 
 In this paper we propose three refinements over the original construction. Our {\em main contributions} are:
\begin{enumerate}
    \item The original method assumed that the noises are distributed {\em symmetrically} about zero. Here, we replace this assumption with a {\em distributional invariance} principle. As the i.i.d.\ nature of the noises already satisfy a distributional invariance, i.e., to {\em permutations}, this allows discarding the symmetricity assumption. On the other hand, if we know that the distriubions are symmetric, it can still be exploited by incorporating that knowledge in the applied {\em transformation group}.
    \smallskip
    \item An important part of the original method is that we need to {\em estimate} the $\mathcal{L}^2$ norm of the target band-limited function. Here, we suggest a more efficient way to estimate this norm by tightening the {\em constraints} of the underlying {\em convex optimization} problem.
    \smallskip
    \item Finally, the constraint tightening idea is also applied to enhance the construction of the {\em confidence intervals} at each input, which results in less conservative region estimates. The new method comes with the same types of guarantees as the original method has.
\end{enumerate}
The refined construction is supported by theoretical guarantees as well as several numerical experiments.

\section{Theoretical background: Reproducing Kernels and Paley-Wiener Spaces}
\vspace*{-1mm}
Kernel methods are based on the concept of Reproducing Kernel Hilbert Spaces (RKHSs) and have a wide range of applications in machine learning, system identification and statistics \citep{berlinet2004reproducing}. A core part of their popularity is the {\em representer theorem}, which states that a regression problem in an infinite dimensional RKHS can be traced back to a finite dimensional problem.

\subsection{Reproducing Kernel Hilbert Spaces}
\vspace*{-2mm}
Let $\mathcal{H}$ be a Hilbert space of functions, $f: \mathbb{X} \to \mathbb{R}$, with an inner product $\langle\cdot,\cdot\rangle_\mathcal{H}$. If every Dirac (linear) functional, which evaluates functions at a point, $\delta_z: f \to f(z)$, is continuous for all $z \in \mathbb{X}$ at any given $f \in \mathcal{H}$, then $\mathcal{H}$ is called a \textit{Reproducing Kernel Hilbert Space (RKHS)}.

Every RKHS has a unique {\em kernel}, $k: \mathbb{X} \times \mathbb{X} \to \mathbb{R}$, which is a symmetric and positive definite function with the so-called \textit{reproducing property}, $\langle k(\cdot,z),f \rangle_\mathcal{H} = f(z),$ for each $z \in \mathbb{X}$ and $f \in \RKHS$. A consequence of this is that for any given $z,s \in \mathbb{X}$, we also have
$k(z,s)=\langle k(\cdot,z),k(\cdot,s) \rangle_\mathcal{H}.$

According to the Moore-Aronszajn theorem, it holds true, as well, that for every positive definite and symmetric function, there uniquely exists an RKHS for which it is its reproducing kernel \citep{berlinet2004reproducing}.

The {\em Gram matrix} of kernel $k$ with respect to given inputs $x_1,x_2,...,x_n$ is $K_{i,j} := k(x_i,x_j)$ for all $i,j \in \{1,2,...,n\}$. Note that matrix $K \in \mathbb{R}^{n \times n}$ is always positive semidefinite. A kernel is called \textit{strictly} positive definite, if its Gram matrix is positive definite for all distinct $\{x_i\}$ inputs.

\subsection{Paley-Wiener Spaces}
\vspace*{-2mm}
A \textit{Paley-Wiener space}, $\mathcal{H}$, is a subspace of $\mathcal{L}^2 (\mathbb{R})$, where for each $\varphi \in \mathcal{H}$ the {\em support} of the {\em Fourier transform} of $\varphi$ is included in a given interval $[-\eta, \eta\hspace{0.3mm}]$, where $\eta > 0$ is a hyper-paramter. By denoting the Fourier transform of $\varphi$ by $\hat{\varphi}$, this means that \citep{iosevich2015exponential}:
$$
\varphi(\xi)\, = \int_{-\eta}^{\eta} e^{2\pi i  x\xi} \hat{\varphi}(x)\hspace{0.5mm} \mbox{d}x,
$$
thus a Paley-Wiener space contains {\em band-limited} functions.

Since $\mathcal{H}$ is a subspace of $\mathcal{L}^2$, it inherits its inner product.
A Paley-Wiener space is also an RKHS with kernel\vspace{1mm}
$$k(z,s) \,\defeq\, \frac{\sin({\eta(z-s)})}{\pi(z-s)},\vspace{1mm}$$
where $(z,s) \in \mathbb{R}^2$ with $s \neq z$, and $k(s,s) \defeq \eta / \pi$.
From now on, we work with the \textit{Paley-Wiener kernel} define above.

\section{Problem setting}
\vspace*{-2mm}
Let $(x_1,y_1), (x_2,y_2), ..., (x_n,y_n)$ be a (finite) i.i.d.\ sample of input-output pairs having an unknown $\mathbb{P}_{X,Y}$ joint probability distribution, where $x_k$ and $y_k$ are real-valued, and $\mathbb{E}[\hspace{0.3mm}y_k^2\hspace{0.3mm}] < \infty$. For all $k \in [\hspace{0.3mm}n\hspace{0.3mm}] \defeq \{1,...,n\}$, we have\vspace{0.5mm}
$$
y_k = f_*(x_k)+\varepsilon_k,
\vspace{0.5mm}
$$
where $\{\varepsilon_k\}$ are the noise terms on the \textit{true} or {\em target} function $f_*$ with $\mathbb{E}[\hspace{0.3mm}\varepsilon_k\hspace{0.3mm}] = 0$. Note that $f_*$ can be written as $f_*(x) = \mathbb{E} [\hspace{0.5mm}Y\hspace{0.3mm} |\hspace{0.3mm} X = x\hspace{0.3mm}],$ known as the \textit{regression function}, where $(X,Y)$ is a random vector with distribution $\mathbb{P}_{X,Y}.$

\subsection{Objectives}
\vspace*{-2mm}
Our primary goal is to construct {\em simultaneous confidence bands} for the unknown $f_*$ function, which bands have {\em distribution-free} guarantees with (user-chosen) confidence probabilities for {\em finite} (possibly small) sample sizes. 

More precisely, we aim at constructing a function $I: \mathcal{D} \to \mathbb{R} \times \mathbb{R}$, where $\mathcal{D}$ is the \textit{support} of the input distribution, such that $I(x) = (I_1(x), I_2(x))$ specifies the {\em endpoints} of an interval estimate for the unknown $f_*(x)$, for every $x \in \mathcal{D}$,
\vspace{1mm}
$$\nu(I)\, \defeq\, \mathbb{P}\big(\,\forall x \in \mathcal{D} : I_1(x) \leq f_*(x) \leq I_2(x)\big) \geq 1-\delta,\vspace{1mm}$$
where $\delta \in (0,1)$ is a user-chosen probability, often referred to as \textit{risk}. The quantity $\nu(I)$ is called the \textit{reliability} of the confidence band. By introducing the notation\vspace{1mm}
$$\mathcal{I}\, \defeq\, \big\{ (x,y) \in \mathcal{D}  \times \mathbb{R}: y \in [I_1(x), I_2(x)] \big\},\vspace{1mm}$$
the reliability is then $\nu(I) = \mathbb{P}(\mbox{graph}_{\mathcal{D}} (f_*) \subseteq \mathcal{I}),$ where we used the definition $\mbox{graph}_{\mathcal{D}}(f_*) \defeq \{ (x, f_*(x)): x \in \mathcal{D}  \}$. 
\subsection{Assumptions}
\vspace*{-2mm}

The main assumptions of the {\em original construction} are:
\smallskip

\begin{assumption}
\label{A0} 
{\em The dataset, namely $(x_1, y_1), \dots, (x_n, y_n) \in \mathbb{R} \times \mathbb{R}$, is an i.i.d.\ sample of input-output pairs; $\mathbb{E}[\hspace{0.3mm}y^2_0\hspace{0.3mm}] < \infty$.}
\end{assumption}
\smallskip

\begin{assumption}
\label{A1} {\em Each $\varepsilon_k \doteq y_k - f_*(x_k)$, for $k \in [\hspace{0.3mm}n\hspace{0.3mm}]$, has a symmetric probability distribution about zero and $\mathbb{E}[\hspace{0.3mm}\varepsilon_k\hspace{0.3mm}]=0$. Random variables $x_k$ and $\varepsilon_k$ are independent for all $k \in [\hspace{0.3mm}n\hspace{0.3mm}]$.}
\end{assumption}
\smallskip

\begin{assumption}
\label{A2} {\em The inputs, $\{x_k\}$, are distributed uniformly on $[\hspace{0.4mm}0, 1\hspace{0.2mm}]$.}
\end{assumption}
\smallskip

\begin{assumption}
\label{A3}
{\em The function $f_*$ is from a Paley-Wiener space, $\forall\,  x\in[\hspace{0.4mm}0, 1\hspace{0.2mm}]: |f_*(x)| \leq 1$; and
$f_*$ is almost time-limited to $[\hspace{0.4mm}0, 1\hspace{0.3mm}]:$
$$\int_{\mathbb{R}} f^2_*(x)\,\mathbb{I}(x \notin  [\hspace{0.4mm}0, 1\hspace{0.2mm}]) \: \mathrm{d}x \, \leq \, \delta_0,$$ 
where $\mathbb{I}(\cdot)$ is an indicator and $\delta_0 > 0$ is a
constant.}
\end{assumption}

These assumptions are rather mild, may be apart from A3, and are discussed in detail in \citep{csaji2022nonparametric}. A3 basically means that the {\em distribution of the inputs} must be {\em known}. Although uniform inputs are assumed, the case of many other classes of distributions can be traced back to this assumption \citep{csaji2022nonparametric}.

\section{High-Level Overview of the\\
Confidence Band Construction}
\vspace*{-2mm}
\label{sec:highlevel}
In this section we briefly overview the main ideas and building blocks of the confidence band construction proposed in \citep{csaji2022nonparametric}. The improvements suggested in this paper do not change this high-level picture, they only refine how the actual tasks are carried out.

First, let us recall that for a dataset $\{(x_k,z_k)\}$, where the inputs $\{x_k\}$ are distinct (which happens with probability one under A\ref{A2}), the element from $\mathcal{H}$ which {\em interpolates} every output $z_k$ at the given input $x_k$ and has the {\em smallest kernel norm} \citep{berlinet2004reproducing}, that is\vspace{1mm}
$$\bar{f} \,\defeq\, \argmin \big\{\,\|\hspace{0.3mm}f\hspace{0.4mm}\|_{\mathcal{H}} : f \in \mathcal{H}\hspace{1.5mm} \&\hspace{1.5mm} \forall\hspace{0.3mm} k \in [n]: f(x_k) =\, z_k   \,  \big\},\vspace{1mm}$$
exists and takes the following form for all input $x \in \mathbb{X}$:
$$\bar{f}(x) = \sum_{k=1}^n \hat{\alpha}_k k(x,x_k),$$
where the weights are $\hat{\alpha} = K^{-1} z$ with $z \defeq (z_1,...,z_n)\tr$ and $\hat{\alpha} \defeq (\hat{\alpha}_1,...,\hat{\alpha}_n)$. Note that under our assumptions, namely 
A\ref{A2} and A\ref{A3}, this Gram matrix is almost surely invertible.

Then, the reproducing property implies $\norm{f_*}_\mathcal{H}^2 = \hat{\alpha}\tr K \hat{\alpha}.$ To help building intuitions, we can also recall that a kernel norm can be seen as a {\em measure of smoothness}. In  Paley-Wiener spaces the kernel norm coincides with the $\mathcal{L}^2$ norm, which is used to measure the {\em energy} of functions, as well. 

The main building blocks of the approach are:

\begin{enumerate}
    \item[(i)] First, we need to construct a guaranteed {\em simultaneous confidence region}, $\Theta$, for some of true (noiseless) outputs of the target function at the {\em observed} inputs. Namely, we need to construct a set that stochastically guarantees to contain $f_*(x_k)$, for $k = 1,\dots, d$, where $d \leq n$ is user-chosen (since the data is i.i.d., it is w.l.o.g.\ that we choose the first $d$). 
    This is a nontrivial task, nonetheless it is ``easier'' than constructing a confidence band for the  {\em whole} function. For this step, we build on the results of \citep{csaji2019distribution}.
    \smallskip
    \item[(ii)] Using the confidence set $\Theta \subseteq \mathbb{R}^d$ for the true values of $f_*$ at some observed inputs, constructed in step (i), we calculate a high probability {\em upper bound}, $\tau$, for the kernel {\em norm} (square) of the true function.
    \smallskip
    \item[(iii)] Then, for each {\em input query} point $x_0 \in \mathcal{D}$, we can construct a {\em confidence interval} for $f_*(x_0)$ as follows. We keep a candidate value $z_0$ in the confidence region {\em if and only if} there is a $z = (z_1, \dots, z_d)\tr \in \Theta$, such that the {\em minimum norm interpolation} of the dataset $\{(x_k,z_k)\}_{k=1}^{d} \cup \{(x_0,z_0) \}$ has a norm (square) less than or equal to $\tau$ (i.e., our upper bound for $\norm{f_*}_\mathcal{H}^2$).
\end{enumerate}

In order to make this approach applicable, apart from the method in (i), we need a way to guarantee an upper bound for the norm (square) of the true function. Besides that, we also need  to give an efficient method to compute the endpoints of the confidence intervals for every $x_0 \in \mathcal{D}$.

\section{Gradient-Perturbation Methods}
\vspace*{-2mm}

We use the {\em Kernel Gradient-Perturbation} (KGP) method \citep{csaji2019distribution} for step (i).
KGP is a generalization of the {\em Sign-Perturbed Sums} (SPS) method \citep{csaji2014sign}, hence we start with a brief overview of SPS. 
\subsection{Sign-Perturbed Sums}
\vspace*{-2mm}
The standard SPS method can construct {\em exact}, {\em nonasymptotic} and {\em distribution-free} confidence regions for the true parameters of {\em linear regression} problems, such as
\begin{equation}
\label{LS-task}
y_k\, \defeq\, \varphi_k\tr \theta^* + \varepsilon_k,
\vspace{0.3mm}
\end{equation}
for $k = 1,\dots, n$, where $\theta^* \in \mathbb{R}^d$ is the "true" parameter.

The SPS construction can be best understood as a way to test the following {\em hypothesis}: $\theta = \theta^*$. If it holds, one can compute the exact realization of the noise terms $\varepsilon \defeq (\varepsilon_1, ..., \varepsilon_n)\tr$ by ``inverting'' the system in \eqref{LS-task}. Then, it builds several {\em perturbed datasets} based on the (estimated) noise terms and using a {\em symmetricity} assumption (A\ref{A1}). The hypothesis is accepted if the new datasets are ``similar'' to the original, which is decided based on a {\em rank-test}.

In the core of algorithm, there are {\em evaluation functions}, 
\begin{equation*}
\label{evalfun}
Z_i(\theta)  \, \defeq \, \|\,\Psi^{\nicefrac{1}{2}}\Phi\tr G_i \big( y - \Phi \theta \big)\,\|^2_2,
\end{equation*}
for $i \in \{0, 1, \dots, m-1\}$, where $\Phi \defeq [\varphi_1,...,\varphi_n]^T, \Psi = (\Phi\tr\Phi)^{-1}$, $m > 0$ is a user-chosen integer, $G_0 \,\doteq\, I$, the identity matrix, and for $i \neq 0$, $G_i \,\doteq \,\mbox{diag}(\alpha_{i,1}, \dots, \alpha_{i,n})$; $\{\alpha_{i,j}\}$ are i.i.d.\ Rademacher variables (random variables which take values $+1$ and $-1$ with  probability $\nicefrac{1}{2}$ each); and $\mbox{diag}(\cdot)$ builds a diagonal matrix from the argument.

For the case $\theta = \theta^*$, we have $y - \Phi \theta = \varepsilon$, then
\begin{equation*}
Z_0(\theta^*) \, = \, \|\,\Psi^{\nicefrac{1}{2}}\Phi\tr \varepsilon \,\|^2_2 \,\,{\,\,\buildrel d \over =\,}\,\, \|\,\Psi^{\nicefrac{1}{2}}
\Phi\tr G_i\, \varepsilon \,\|^2_2  \,= \, Z_i(\theta^*), 
\end{equation*}
for $i=1,...,m-1$, where ``$\eqind$'' denotes equality in distribution. These variables are not independent, however, they are {\em exchangeable} \citep{csaji2014sign}. On the other hand, as $\norm{\theta - \theta^*}_2$ increases, the chance that $Z_0(\theta)$ dominates the other (perturbed) $\{Z_i(\theta)\}_{i\neq 0}$ variables increases, as well.

The normalized rank of $Z_0 (\theta)$ is defined as
\begin{equation*}
\mathcal{R}(\theta) \,\doteq \, \frac{1}{m}\, \bigg[ 1 +  \sum\nolimits_{i = 1}^{m-1} \mathbb{I}\left( Z_0(\theta) \prec Z_i(\theta) \right) \bigg],
\end{equation*}
where $\mathbb{I}(\cdot)$ is an indicator function (the value it takes is $1$ if its argument is true and $0$ otherwise), and ``$\prec$'' is the usual ``$<$'' with random tie-breaking \citep{csaji2014sign}.

Any rational target confidence probability $p \in (0,1)$ can be written in the form of $p = 1 - q / m$, where $0 < q < m$ are integers. The SPS method will accept the hypothesis $\theta = \theta^*$ if $\mathcal{R}(\theta) \leq p$, and rejects it otherwise. Hence, the SPS confidence region is defined as:
\begin{equation*}
\widehat{\Theta}_{p} \,\doteq \, \left\{ \, \theta \in \mathbb{R}^d : \mathcal{R}(\theta) \leq p \, \right\}\!.
\end{equation*}
It can be proved that $\mathbb{P}(\theta^* \in \widehat{\Theta}_{p})\, =\, p$, i.e., these regions have {\em exact} confidence probabilities \citep{csaji2014sign}.

\subsection{Ellipsoidal Outer-Approximation of SPS Regions}
\vspace*{-2mm}
We can construct ellipsoidal outer approximations for the SPS confidence regions \citep{csaji2014sign} taking the form
\begin{equation*}
\widehat{\Theta}_p \, \subseteq \,\widetilde{\Theta}_p\, \defeq\, \big\{\, \theta \in \mathbb{R}^d\, :\, (\theta-\hat{\theta}_n)^\mathrm{T}R_n(\theta-\hat{\theta}_n)\leq 
\gamma^*\, \big\},
\end{equation*}
where $\hat{\theta}_n$ is the LS estimate, $R_n \defeq \nicefrac{1}{n}\,\Phi \tr \Phi$ and the radius, $\gamma^*$, is the $q$\,th largest of the $\{\gamma_i\}_{i=1}^{m-1}$ values defined by
$$
\gamma_i \, \defeq  \max_{\{\theta: Z_0(\theta) \leq Z_i(\theta)\}}{Z_i(\theta)}.
$$
Unfortunately, the optimization problems above are {\em not} convex. Nevertheless, it can be proven by building on duality theory that the following (convex) {\em semi-definite} problem has the same optimal value \citep{csaji2014sign}:
\begin{equation}
\begin{aligned}
\label{SPSopt}
\mathrm{minimize}\quad & \gamma  \\
\mbox{subject to}\quad &
\lambda\geq 0  \\
 &
\left[\begin{array}{cc} -I+\lambda A_{i} & \lambda b_{i} \\ \lambda b_{i}^\mathrm{T} & \lambda c_{i} + \gamma\end{array}\right]\succeq 0, 
\end{aligned}
\vspace{1mm}
\end{equation}
where ``$\succeq$'' denotes p.s.d.\ ordering, and $A_{i}$, $b_{i}$ and $c_{i}$ are\vspace{-3mm}
\begin{eqnarray*}
A_{i}&\defeq&I-R_n^{-\frac{1}{2}}Q_{i}R_n^{-1}Q_{i}R_n^{{-\frac{1}{2}\mathrm{T}}},\\
b_{i}&\defeq&R^{-\frac{1}{2}}_n Q_{i} R_n^{-1}(\psi_i-Q_i\hat{\theta}_n),\\
c_{i}&\defeq&-\psi_i^\mathrm{T} R_n^{-1}\psi_i+2\hat{\theta}^\mathrm{T}_n Q_i R_n^{-1}\psi_i- \hat{\theta}^\mathrm{T}_n Q_i R_n^{-1} Q_i \hat{\theta}_n,\\[-5mm]
\end{eqnarray*}
where matrix $Q_i$ and vector $\psi_i$ take the form
\begin{equation*}
\label{SPSperm}
Q_{i} \defeq \frac{1}{n}\Phi\tr G_i\Phi, \qquad \text{and}\qquad \psi_{i} \defeq \frac{1}{n} \Phi\tr G_i\, y,
\end{equation*}
where $y \defeq (y_1, \dots, y_n)\tr$ is the vector of outputs.

Due to its construction, we have $\mathbb{P}(\theta^* \in \widetilde{\Theta}_p)\, \geq\, p$.

\subsection{Kernel Gradient Perturbation}
\vspace*{-2mm}
To construct a confidence ellipsoid for the first $d$ function values of $f^*$, i.e., step (i) of the algorithm, we apply the KGP method \citep{csaji2019distribution}, an extension of SPS. 

KGP builds confidence regions for {\em ideal} representations. A representation $f \in \CH$ is called ideal w.r.t.\ $\{x_k\}_{k=1}^{d}$, if it has the property that $f(x_k) = f_*(x_k)$, for all $k \in [\hspace{0.3mm}d\hspace{0.5mm}]$.

Let us consider the following problem, for a given $\lambda \geq 0$:\vspace{0.5mm}
\begin{equation*}
\text{minimize}\;\;(y - \Ker_1\hspace{0.2mm} \theta)\tr (y - \Ker_1\hspace{0.2mm} \theta) \,+\, \lambda\, \theta\tr \hspace{-0.3mm}\Ker_2\hspace{0.2mm} \theta,
\vspace{0.5mm}
\end{equation*}
where $K_1 \in \RR^{n \times d}$ is the Gram matrix $K$ having the last $n-d$ columns removed, and $K_2\in \RR^{d \times d}$ is $K_1$ having the last $n-d$ rows removed. 
This can be reformulated as a {\em least squares} problem, $\|\hspace{0.3mm}{\newpart v} \,-\, \Phi\hspace{0.2mm} \theta\hspace{0.3mm}\|^2$, by using
\vspace{0.5mm}
\begin{equation*}
\Phi\, =\, \left[ 
\begin{array}{c}
\, \Ker_1\, \\[1mm]
\sqrt{\lambda}\, \Ker_2^{\!\frac{1}{2}}
\end{array}  
\right]\!,\qquad
{\newpart v} \,=\, \left[ 
\begin{array}{c}\,
\, y\, \\[1mm]
\;0\;
\end{array}  
\right]\!,
\vspace{-0.5mm}
\end{equation*}
where $\Ker_2^{\!\frac{1}{2}}$ denotes the principal, non-negative square root of $\Ker_2$, which exists as $K_2$ is positive semi-definite.

We search for $\tilde{\theta} \in \RR^d$ {\em ideal vector}, such that for every $k \in [\hspace{0.3mm}d\hspace{0.5mm}]$, we have $(K_1 \tilde{\theta})(k)=  f_*(x_k)$. We can apply the outer approximation approach of SPS to the reformulated problem to build a guaranteed confidence ellipsoid for ideal vector $\tilde{\theta}$.
Note that only the first $d$ residuals should be perturbed, when SPS is applied, as we can only reconstruct the first $d$ noise variables \citep{csaji2022nonparametric}.
\subsection{First Refinement: Distributional Invariance}
\vspace*{-2mm}
\label{sec:first_ref}
The original confidence region construction assumed symmetric noises (see A\ref{A1}), as it mainly applied SPS, but this assumption can be relaxed. It is in fact enough if we assume a {\em distributional invariance} for the noises, that is
\smallskip
\begin{assumption}
\label{A1'} {\em Each noise term has zero mean, $\mathbb{E}\big[\varepsilon_k\big]=0$, 
variables $x_k$ and $\varepsilon_k$ are independent, for $k \in [n]$, and for a compact\vspace{-0.6mm} matrix group,  $\mathcal{G} \subseteq \mathbb{R}^{n \times n}$, it holds that $\forall\, G \in \mathcal{G}: G\hspace{0.5mm} \varepsilon \eqind \varepsilon$.}
\end{assumption}

The stochastic guarantees of KGP methods remain valid under this assumption, as well \citep{csaji2019distribution}.

Symmetric noises are special cases of A\ref{A1'}, as we can use the group of diagonal matrices which contain only $+1$ and $-1$ values as $\mathcal{G}$. Other examples are, e.g., if the noises are {\em exchangeable}, the group of {\em permutation} matrices satisfy this property. As A\ref{A0} guarantees exchangeability, the symmetricity assumption is not needed anymore for the permutation variant. For SPS, the permutation variant was originally proposed in \citep{kolumban2015perturbed}.

Because of A\ref{A0}, the default choice for the refined confidence band method is the group of permutation matrices. 

The construction of SPS then can be recasted with any matrix group that guarantees distributional invariance. For example, if we use the group of permutations, the construcion of ellipsoidal outer approximation remains the same, only matrix $Q_i$ and vector $\psi_i$ changes to
\begin{equation*}
Q_{i} \defeq \frac{1}{n}\Phi\tr P_i\Phi, \qquad \text{and}\qquad \psi_{i} \defeq \frac{1}{n} \Phi\tr P_i\, y,
\end{equation*}
where $P_i$ is a random (uniform) permutation matrix.

As in the case of sign-changes, if we apply random permutations when we construct a confidence ellipsoid for the ideal vector $\tilde{\theta}$, we should only perturb the indices of the first $d$ residuals, as we can only reconstruct the first $d$ noise terms. Therefore, we should use matrices 
$$G_i \,=\, \left[\begin{array}{cc} P_i & 0 \\ 0 & I_{n-d}\end{array}\right]\!,$$
where $P_i \in \mathbb{R}^{d \times d}$ is a random permutation matrix, and $I_{n-d} \in \mathbb{R}^{(n-d) \times (n-d)}$ is the identity matrix.

\section{Upper Bound for the Kernel Norm }
\vspace*{-2mm}
The original construction estimated $\norm{f_*}_{\CH}^2$ as follows. Let us define $\varphi_k \defeq (k(x_1,x_k), \dots, k(x_n,x_k))\tr$, we know that $f_*(x_k) = \varphi_k\tr\tilde{\theta}$, for $k \in [\hspace{0.3mm}d\hspace{0.5mm}]$, where $\tilde{\theta}$ is the (unknown) parameter vector of the ideal representation. We saw that for any (rational) probability $\beta \in (0,1)$ we can construct a confidence ellipsoid, $\widetilde{\Theta}_{\beta}$, such that it contains $\tilde{\theta}$ with probability at least $1-\beta$. Then, we can construct (probabilistic) upper and lower bounds of $f_*(x_k)$ by maximizing and minimizing $\varphi_k\tr\theta$, for $\theta \in \widehat{\Theta}_{\beta}$. Let us introduce
\begin{equation*}
\nu_k \defeq \min_{\theta \in \widetilde{\Theta}_{\beta}} \varphi_k\tr \theta \qquad \text{and} \qquad \mu_k \defeq \max_{\theta \in \widetilde{\Theta}_{\beta}} \varphi_k\tr \theta,
\end{equation*}
for all $k \in [\hspace{0.3mm}d\hspace{0.5mm}]$, which (convex) problems have analytical solutions \citep{csaji2022nonparametric}. Then, the intervals $[\nu_k, \mu_k]$, for $k \in [\hspace{0.3mm}d\hspace{0.5mm}]$, are simultaneous confidence intervals for the first $d$ functions values, $f_*(x_k)$, for $k \in [\hspace{0.3mm}d\hspace{0.5mm}]$. That is\vspace{1mm}
\begin{equation*}
\mathbb{P}\big(\hspace{0.3mm} \forall \hspace{0.3mm}k \in [\hspace{0.3mm}d\hspace{0.5mm}]: f_*(x_k) \in [\hspace{0.3mm}\nu_k, \mu_k\hspace{0.3mm}]\hspace{0.3mm}\big)\, \geq\, 1 - \beta.
\vspace{1mm}
\end{equation*}
Using these intervals, an upper bound for $\norm{f_*}_{\CH}^2$ is
\vspace{-0.5mm}
\begin{equation*}
\tau \, \defeq\, \frac{1}{d} \sum_{k=1}^{d} \max\{\nu^2_k,\mu^2_k \} + \sqrt{\frac{\ln(\alpha)}{-2d}} + 
\delta_0,
\end{equation*}
where $\alpha \in (0,1)$ is a risk probability. This bound construction guarantees that \citep{csaji2022nonparametric}\vspace{1mm}
$$\mathbb{P}\big(\norm{f_*}_{\CH}^2 \leq \tau \hspace{0.3mm}\big)\,  \geq \,1-\alpha-\beta.\vspace{1mm} $$
A fundamental property which made this bound construction possible is that the kernel norm of a Paley-Wiener space coincides with the well-known $\mathcal{L}^2$ norm.

\subsection{Second Refinement: Improved Norm Bound}
\vspace*{-2mm}
In this section, we present a more efficient way to construct an upper bound for $\norm{f_*}_{\CH}^2$. The issue with the original construction is that the intervals $[\nu_k, \mu_k]$, for $k \in [\hspace{0.3mm}d\hspace{0.5mm}]$, are constructed independently, as if choosing a function value at an input could not influence the choice of function values at other inputs. This might lead to conservative results. 

Assume for simplicity that $\lambda = 0$. Then $\Phi = K_1$ and the ellipsoidal outer approximation takes the following form 
\vspace{1mm}
\begin{equation}
\label{ellipsoid-normal-form}
\widehat{\Theta}_{\beta} \; \defeq \; \big\{\, \theta \in \mathbb{R}^ n\, :\, (\theta-\widehat{\theta})^\mathrm{T}(\nicefrac{1}{n})
\,K_1\tr K_1\hspace{0.3mm}(\theta-\widehat{\theta})\,\leq\, \gamma^* \, \big\},
\vspace{1mm}
\end{equation}
where $\widehat{\theta}$ is the LS estimate and  $\beta$ is a risk probability.

After dividing both sides of \eqref{ellipsoid-normal-form} by $\gamma^*$, and by introducing $H \defeq \frac{1}{n \gamma^*} K_1\tr K_1$, the confidence ellipsoid becomes
\begin{equation}
\label{ellipsoid-form-1}
\widehat{\Theta}_{\beta} \; \defeq \; \big\{\, \theta \in \mathbb{R}^n\, :\, (\theta-\widehat{\theta})^\mathrm{T}H\hspace{0.3mm}(\theta-\widehat{\theta})\,\leq\, 1 \, \big\},
\vspace{1mm}
\end{equation}
This ellipsoid contains (with high probability) the {\em coefficients} of the ideal representation. The function values of the ideal representation can be calculated using the matrix $K_2$. Hence, in order to get a confidence ellipsoid for the {\em function values} at the first $d$ inputs, we need to transform ellipsoid \eqref{ellipsoid-form-1} by $K_2$. By multiplying both sides of \eqref{ellipsoid-form-1} by $K_2$, the Hessian of the ellipsoid becomes $K_2^{-1} H K_2^{-1}$, and the center will be $K_2\hat{\theta}$. Finally, we arrived at
\vspace{1mm}
\begin{equation*}
\mathcal{Z} \; \defeq \; \big\{\, z \in \mathbb{R}^d\, :\, (z-K_2\widehat{\theta})^\mathrm{T}K_2^{-1}H K_2^{-1}\hspace{0.3mm}(z-K_2\widehat{\theta})\,\leq\, 1 \, \big\},
\vspace{1mm}
\end{equation*}
which, by construction, has the property that
\vspace{1mm}
\begin{equation}
\label{conf_ellips_fs}
\mathbb{P}\big(\, (f_*(x_1), \dots, f_*(x_d))\tr \in \mathcal{Z}\,\big) \, \geq \, 1-\beta.
\vspace{1mm}
\end{equation}
With this, we can provide an improved upper bound for the norm. Let us denote $z \defeq (z_1, ..., z_d).$ Instead of using the absolute maximum of every single $d$ data points, we can solve an optimization problem with respect to $\mathcal{Z}$ as:
\begin{equation}
\label{noisy-norm-max-rewritten}
\begin{split}
\mbox{minimize} \; -\frac{1}{d}\, \|z\|^2  \qquad
\mbox{subject to} \;\, z \in \mathcal{Z}.\\[1mm]
\end{split}
\end{equation}
This problem is not convex, but thanks to strong duality, we can solve the dual problem instead \citep{boyd2004convex}.

By introducing  $A_1  \defeq K_2^{-1} H K_2^{-1}$, $b_1\tr \defeq - \widehat{\theta} \tr H K_2^{-1}$, and $c_1 \defeq \widehat{\theta}^\mathrm{T} H \widehat{\theta}-1$, the constraint of \eqref{noisy-norm-max-rewritten} can be written as\vspace{1mm}
\begin{equation*}
z^\mathrm{T} A_1 z + 2 \, b_1\tr z + c_1 \,\leq\, 0.
\vspace{1mm}
\end{equation*}
With this notation, we can apply a result from \citep[B.1]{boyd2004convex} about the dual of (even nonconvex) quadratic problems that have only one quadratic constraint to get
\vspace{0.5mm}
\begin{equation}
\begin{aligned}
\label{dual-norm-problem}
\mathrm{maximize}\quad\;\; & \xi  \\
\mbox{subject to }\quad &
\varrho \geq 0 \\
 &
\left[\begin{array}{cc} A_0+\varrho A_1 & \varrho\, b_{1} \\ \varrho\, b_{1}^\mathrm{T} & \varrho\, c_{1} - \xi \end{array}\right]\succeq 0, 
\end{aligned}
\vspace{1mm}
\end{equation}
where $A_0 \defeq -\frac{1}{d} I$ comes from the optimization objective.

Problem \eqref{dual-norm-problem} is always convex and can be computed efficiently. If we denote the optimal solution by $\xi^*$, then the upper bound for the norm square of $f_*$ is the following:\vspace{1mm}
\begin{equation*}
\tau_0 \, \defeq\, \xi^* + \sqrt{\frac{\ln(\alpha)}{-2d}} + 
\delta_0.
\vspace{1mm}
\end{equation*}
It could be shown that the refined bound $\tau_0$ comes with the same stochastic guarantees as the original bound $\tau$.

\section{Confidence Intervals at Query Inputs}
\vspace*{-2mm}
The final step of the confidence band construction is that we should be able to provide a confidence interval for any given input {\em query point} $x_0 \in \CD$ with $x_0 \neq x_k$, for $k \in [\hspace{0.3mm}d\hspace{0.5mm}]$.

In the original construction, the boundaries of the confidence intervals are given by {\em two} convex problems:
\vspace{1mm}
\begin{equation}
\label{noisy-opt-min-max}
\begin{split}
\mbox{min\,/\,max} &\quad z_{0} \\[0.5mm]
\mbox{subject to} &\quad (z_0, \dots, z_d)\hspace{0.3mm} {K}_0^{-1} (z_0, \dots, z_d)\tr \leq\, \tau\\[1mm]
&\quad \nu_1 \leq z_1 \leq \mu_1,\; \dots,\; \nu_d \leq z_d \leq \mu_d,\\[1mm]
\end{split}
\end{equation}
where ``min\,/\,max'' means that the problem must be solved as a minimization and also as a maximization; and \vspace{0.8mm}
$${K}_0({i+1},{j+1})\, \defeq \, k(x_i,x_j), \vspace{1mm}$$
is the extended Gram matrix for $i, j = 0,1, \dots ,d$.

The intuition behind this construction was discussed in Section \ref{sec:highlevel}: we should be able to interpolate each possible point in the intervals $[\nu_k, \mu_k]$, for $k \in [\hspace{0.3mm}d\hspace{0.5mm}]$, as well as $z_0$ with a function that has a norm square not bigger then $\tau$. 

This construction guarantees  \citep{csaji2022nonparametric} \vspace{1mm}
$$\mathbb{P}(\, \mathrm{graph}_{\CD}(f_*) \subseteq \mathcal{I}\,) \, \geq \, 1-\alpha - \beta, \vspace{1mm}$$
under the assumptions A\ref{A0}, A\ref{A1}, A\ref{A2}, and A\ref{A3}.
\subsection{Third Refinement: Improved Confidence Intervals}
\vspace*{-2mm}
We have constructed ellipsoid $\mathcal{Z}$ to satisfy property \eqref{conf_ellips_fs}. Using this, we can also refine the confidence interval construction problem(s) presented by \eqref{noisy-opt-min-max}: the box constraints given by the confidence intervals should be replaced by an ellipsoidal constraint given by $\mathcal{Z}$, formally:
\vspace{1mm}
\begin{equation}
\label{noisy-opt-min-max-modified}
\begin{split}
\mbox{min\,/\,max} &\quad z_{0} \\[0.5mm]
\mbox{subject to} &\quad (z_0, \dots, z_d)\hspace{0.3mm} {K}_0^{-1} (z_0, \dots, z_d)\tr \leq\, \tau_0\\[1mm]
&\quad (z_1, ..., z_d) \in \mathcal{Z},\\[1.4mm]
\end{split}
\end{equation}
where we also used the improved norm bound $\tau_0$. These problems are convex, they can be solved efficiently.

\section{Numerical Experiments}
\vspace*{-2mm}
The algorithms were also implemented and tested numerically.
The Paley-Wiener RKHS was used with parameter $\eta = 30$. The ``true'' data-generating function was constructed as follows: first, $20$ random input points $\{\bar{x}_k\}_{k=1}^{20}$ were generated, with uniform distribution on $[\hspace{0.3mm}0,1]$. Then $f_*(x) = \sum_{k=1}^{20} w_k k(x, \bar{x}_k)$ was created, where each $w_k$ had a uniform distribution on $[-1,1]$. The function was normalized, in case its maximum value exceeded $1$.

\subsection{Confidence Bands for Non-Symmetric Noises}
\vspace*{-2mm}
In the non-symmetric case, we implemented the previously introduced, permutation-based approach, and combined it with the refined convex programs, presented in \eqref{dual-norm-problem}
and \eqref{noisy-opt-min-max-modified}, to construct simultaneous confidence bands. 

We generated $n = 300$ noisy observations from $f_*$. The measurement noise had the following distribution: $\varepsilon \sim \exp({\lambda})-\nicefrac{1}{\lambda}$, where our choice of parameter was $\lambda = 0.25$. This distribution also fulfils the criteria given in A\ref{A1'}, since its expected value is $0$, however, it is not symmetric due to the properties of the exponential distribution. We compared our results on different significance levels.

\begin{figure}[!t]
    \centering
	\hspace*{-2mm}	
	\includegraphics[width = 1.02\columnwidth]{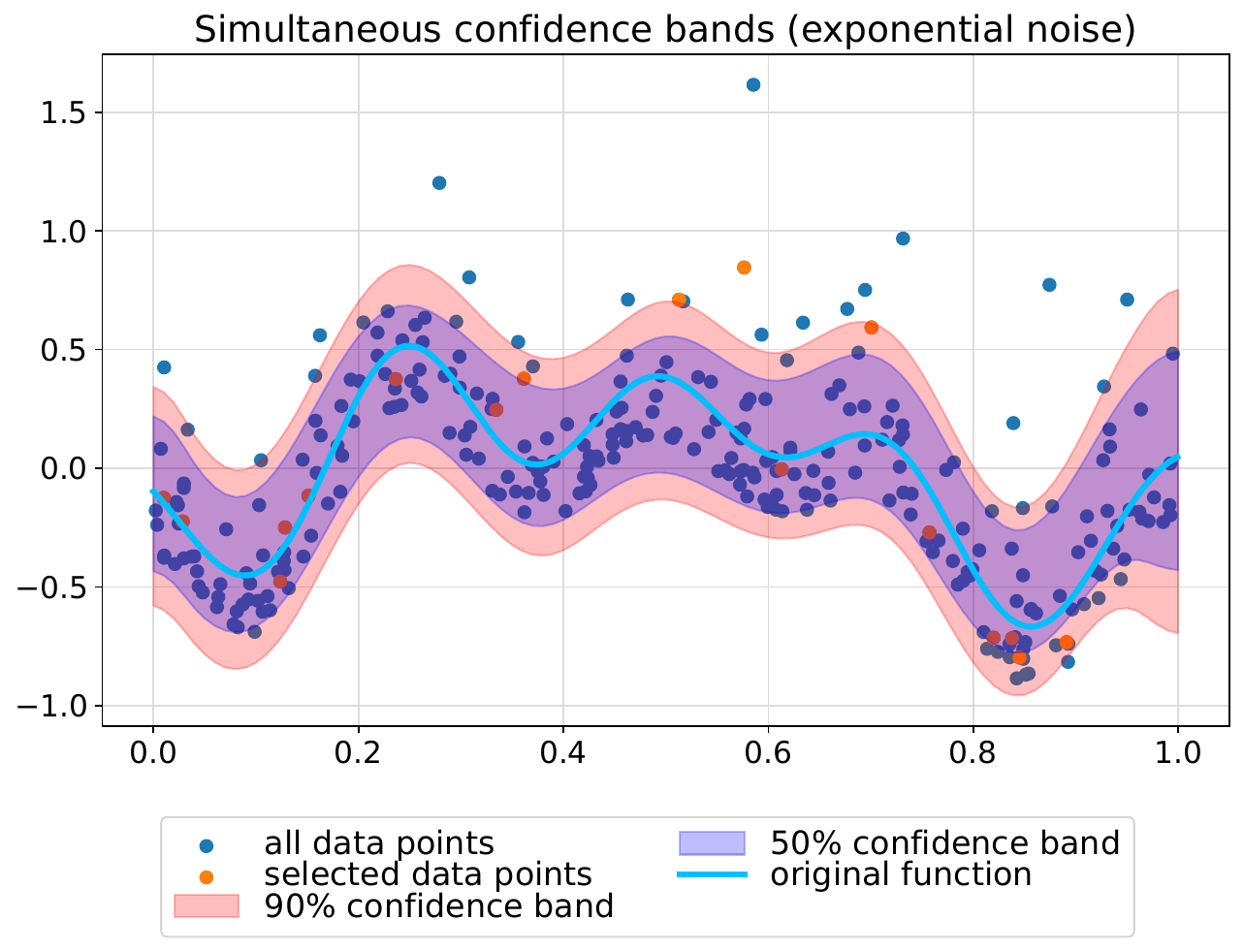} 	
    \caption{Random permutation based simultaneous confidence bands with exponentitally distributed noises; all of the three proposed refinements were used.}
\label{fig:experiment1}
\vspace*{2mm}
\end{figure}

Figure \ref{fig:experiment1} shows that the refined approach leads to informative and adequate simultaneous confidence bands, even when the measurement noise is non-symmetric.

\subsection{Comparing the Original and the Refined Methods}
\vspace*{-2mm}

We also tested our refined convex programs for symmetric noises. In this case, the original sign-perturbation based KGP method was used for constructing the confidence ellipsoid in step (i). The aim was to measure the improvements provided by the reformulated convex programs \eqref{dual-norm-problem}
and \eqref{noisy-opt-min-max-modified} over their original counterparts.

We had $n = 300$ random noisy observations from $f_*$. The measurement noise $\{\varepsilon_k\}$ had Laplace distribution with location $\mu = 0$ and scale $b = 0.25$ parameters.

\begin{figure}[!t]
    \centering
	\hspace*{-2mm}	
    \includegraphics[width = 1.02\columnwidth]{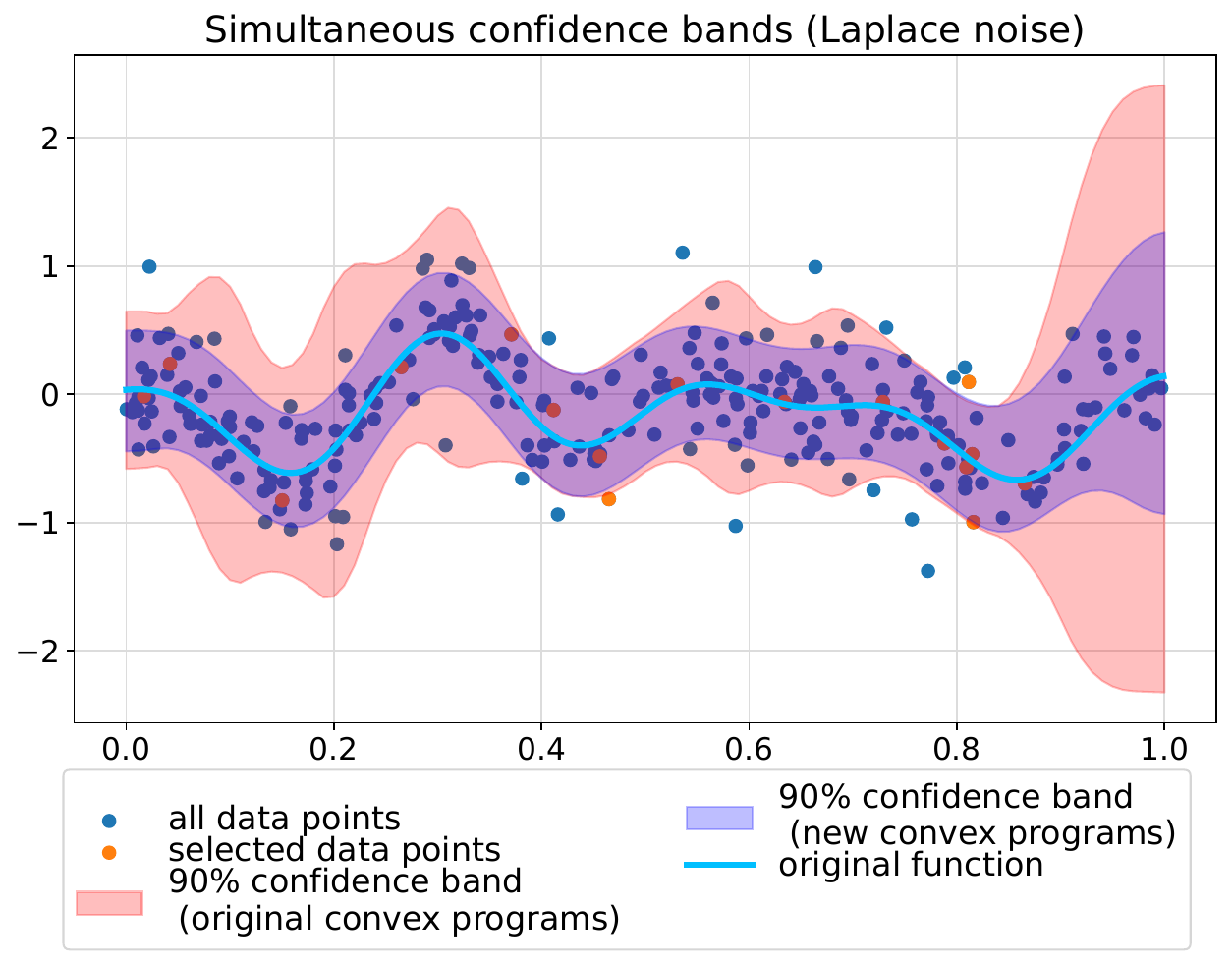} 	
    \caption{Sign-changes based simultaneous confidence bands with Laplace noises; comparing the refined convex programs 
     \eqref{dual-norm-problem} and \eqref{noisy-opt-min-max-modified} with the original ones.}
\label{fig:experiment4}
\vspace*{4mm}
\end{figure}

The experiment presented in Figure \ref{fig:experiment4} 
confirms that the refined construction is more efficient, less conservative.

\begin{remark}
The convex programs in \eqref{dual-norm-problem} and \eqref{noisy-opt-min-max-modified} both include the inverse of the Gramian matrix that may be numerically unstable for certain kernels. There are various ways to handle this, the simplest is to add a tiny constant times the identity matrix to Gramian before inverting it.

\end{remark}

\section{Conclusions}
\vspace{-2mm}
In this paper, we have investigated the problem of constructing nonparametric simultaneous confidence bands with nonasymptotic and distribution-free guarantees. 
The starting point was a recent Paley-Wiener kernel-based construction \citep{csaji2022nonparametric}, for which three improvements were proposed. First, (1) the assumptions about the measurement noises were relaxed, by allowing non-symmetric noises.
Then, (2) the construction of a high-probability upper bound for the 
norm was refined by introducing a convex program to calculate a more efficient bound. Finally, (3) the convex 
programs for building a confidence interval at any given query 
point was refined by replacing the 
box constraints with 
an ellipsoidal one.
\bibliography{references}           %
\end{document}